\documentclass[10pt,twocolumn,letterpaper]{article}

\usepackage{wacv}
\usepackage{times}
\usepackage{epsfig}
\usepackage{graphicx}
\usepackage{amsmath}
\usepackage{amssymb}
\usepackage{makecell} 
\usepackage{gensymb} 
\usepackage{enumitem}
\usepackage{subcaption}
\usepackage{float} 
\usepackage{pdfpages}

\wacvfinalcopy 

\def\wacvPaperID{409} 

\ifwacvfinal\fi
\begin{document}

\title{Driving Scene Perception Network: \\
Real-time Joint Detection, Depth Estimation and Semantic Segmentation}


\author{Liangfu Chen \hspace{2cm} Zeng Yang\\
Harman International\\
{\tt\small \{liangfu.chen,zeng.yang\}@harman.com}
\and
Jianjun Ma\\
{}\\
{\tt\small tonymjj@163.com}
\and
Zheng Luo\\
University of Virginia\\
{\tt\small zl5sv@virginia.edu}
}

\maketitle
\ifwacvfinal\thispagestyle{empty}\fi

\begin{abstract}
As the demand for enabling high-level autonomous driving has increased in recent years
and visual perception is one of the critical features to enable fully autonomous driving,
in this paper, we introduce an efficient approach for simultaneous 
object detection, depth estimation and pixel-level semantic segmentation 
using a shared convolutional architecture. The proposed network model, 
which we named Driving Scene Perception Network (DSPNet), 
uses multi-level feature maps and multi-task learning to improve the accuracy and 
efficiency of object detection, depth estimation and image segmentation tasks 
from a single input image. Hence, the resulting network model uses less than 850 MiB of GPU memory and 
achieves 14.0 fps on NVIDIA GeForce GTX 1080 with a $1024\times{}512$ input image,
and both precision and efficiency have been improved over combination of single tasks.
\end{abstract}

\section{Introduction}

Autonomous driving requires understanding the layout of the surroundings, such as the
distance to vehicles, pedestrians and other obstacles, knowing exactly where the drivable road and sidewalk regions are,
and locating road markings and traffic signs.
To date, we have found no single task model that is capable of simultaneously
performing object detection, depth estimation and pixel-level scene segmentation 
because the bounding box-level detection of drivable road is meaningless 
and the pixel-level segmentation of a group of vehicles is not intuitive in practice.



We divide the visual perception task into object detection, depth estimation and semantic segmentation sub-tasks.
First, the object detection task contains the bundled estimation of both object location and object categories.
Second, we observe that depth estimation is critical and that predicting pixel-level depth for each input image
is expensive and unnecessary, as the distance variation for each detected object is ignored in most cases.
Finally, detection of the road and buildings can be meaningless, even if precise bounding boxes are predicted; 
therefore, we ensure that the network is aware of the road and buildings via pixel-level semantic labeling.
The road segmentation task can be further extended to consider road marking segmentation tasks; 
therefore, lane markings, directional arrow markings and pedestrian crossings can be estimated easily.

\begin{figure}[tbp]
\begin{center}
\includegraphics[width=3.2in]{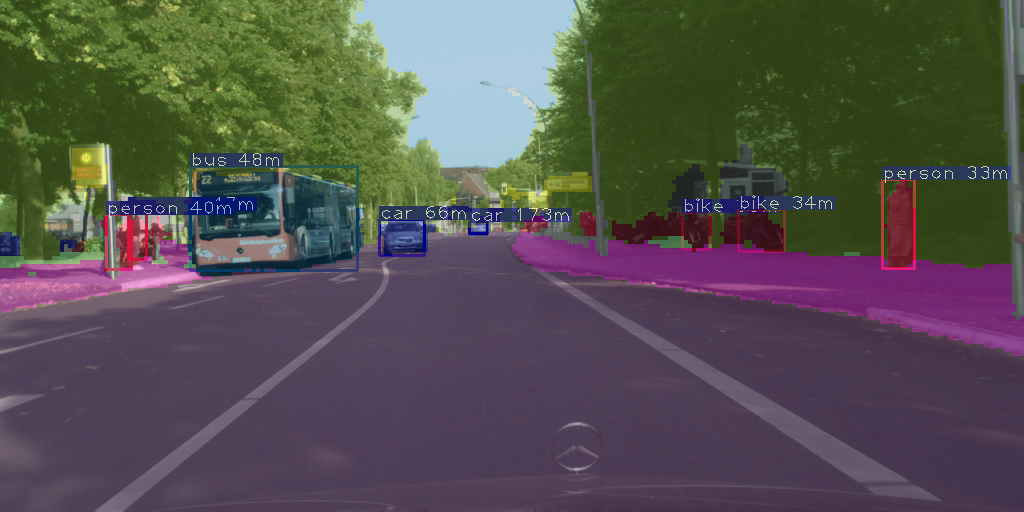} 
\caption{A typical result showing detected objects with estimated distance, 
with the semantic segmentation result overlaid. 
With awareness of the instance-level distance for each detected object and 
pixel-level semantic segmentation of drivable regions, 
a frontal collision avoidance system can be easily implemented.}
\label{fig:cover_image}
\end{center}
\end{figure}


We make choices for the overall architecture design based on several practical considerations: 
1) capability of detecting objects on the road, 
2) capability of estimating the distance to detected objects,
3) capability of segmenting a drivable road,
4) computational efficiency, 
5) small memory footprint, and
6) highly accurate performance of the tasks above.

The ideal design of a multi-task network model that satisfies the critical requirements above should be as follows:
First, the model should be capable of efficiently extracting features using shared convolutional feature maps; 
convolutional neural network models that were pre-trained on a large-scale image classification dataset 
(i.e.,\ ImageNet) are proven to be robust in achieving such a goal.
Second, the model should capable of efficiently decoding the extracted features and predicting bounding boxes, class labels and distance.

To achieve the goal of designing a network model that satisfies the above requirements, 
we use a residual network with 50 weighted layers (a.k.a.\ resnet-50) as a base network for feature extraction;
this method achieves the high gflops-over-accuracy ratio that is needed to perform the image classification task, as stated in \cite{Paszke2016}. 
We also eliminate pixel-level depth prediction and predict the depth at the object-level, which can be easily integrated 
into existing detection frameworks such as Faster RCNN \cite{Ren2015} and SSD \cite{Redmon2016}.

We summarize our contributions as follows:
\begin{itemize}[nosep]
\item \textit{Efficient multi-task inference:} 
An efficient neural network model for simultaneous object detection, depth estimation and 
pixel-level semantic segmentation using a shared convolutional architecture is introduced.
The proposed network model is proven to be more accurate and computationally efficient than a combination of single task models.
\item \textit{Instance-level depth estimation:} 
An efficient approach for instance-level depth estimation that does not require the computation of region proposals is introduced.
To our best knowledge, the proposed model is the first convolutional neural network model 
that can predict instance-level depth without region proposal computations.
\item \textit{Effective training strategy:} 
To support training for a multi-tasking architecture, a data augmentation approach is proposed, and a few architectural decisions are made that balance computational efficiency and prediction accuracy, which aids in the practical application of the multi-task network to automate driving tasks.
\end{itemize}

The remaining sections of the paper are organized as follows: 
Related work is described in section \ref{sec:related_works}; details about the network design are described in section \ref{sec:network_design}; section \ref{sec:details} describes the implementation details of the proposed network model; experimental results are described in section \ref{sec:results}; and the conclusion is summarized in section \ref{sec:conclusion}.

\begin{figure*}
\begin{center}
\includegraphics[width=6.8in]{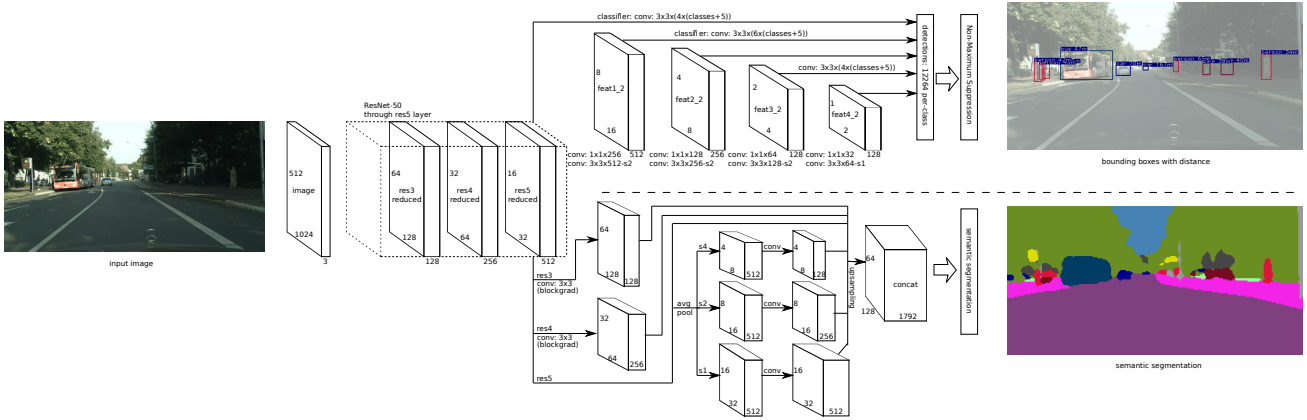}
\caption{The proposed network model performs joint object detection, depth estimation and semantic segmentation tasks with a shared convolutional architecture.}
\label{fig:network_model}
\end{center}
\end{figure*}

\section{Related Works}
\label{sec:related_works}

The ImageNet Large Scale Visual Recognition Challenge \cite{ILSVRC15} provides 
enough labelled image databases for robust image classification over 
a wide range of object categories.
Convolutional neural network models that perform image classification tasks on the ImageNet dataset
have been a common choice in recent years:
AlexNet \cite{Krizhevsky2012}, ZFNet \cite{Zeiler2014}, 
VGGNet \cite{Simonyan2015}, Residual Network \cite{He2015a}
GoogLeNet \cite{Szegedy2014}, Inception V3 \cite{Szegedy2016a}, 
Xception \cite{chollet2016xception} and MobileNet \cite{howard2017mobilenets}.

For more practical application network models, based on the convolutional architecture, detection network models have been further developed.
RCNN \cite{Girshick2016} and its successor Faster-RCNN \cite{Ren2015} 
perform image classification based on region proposals; 
YOLO \cite{Redmon} and SSD \cite{Redmon2016} introduced multiple decoding layers to avoid
the necessary computing region proposals.
Another important reason SSD is favored over Faster-RCNN is 
that visual perception in an autonomous driving system favors overall scene understanding 
over correctness of object classification, for instance, 
an advertisement with a person painted on a bus should be classified as bus instead of pedestrian.

The existing approaches for predicting depth from a single image are either 
pixel-level-based dense prediction \cite{eigen2015predicting,saxena2006learning,liu2016learning},
or traditional approaches that are based on object-level prediction \cite{stein2003vision}.
Pixel-level-based dense prediction approaches commonly borrow ideas from 
Fully Convolutional Networks (FCNs) \cite{Long2015}.
Instead of computing class labels for each pixel, the depth map can be obtained via regression from the feature maps.
A traditional approach-based estimation is not sufficiently robust in the case of complicated urban street scenes. 

For scene segmentation, a classical baseline is settled by FCN \cite{Long2015}, 
which initializes the upsampling kernels via bilinear sample weights 
and predicts an arbitrary image input within feature maps that are down-sampled by 8 times.
SegNet \cite{badrinarayanan2015segnet} and ENet \cite{Paszke2016} are introduced to 
improve the computational efficiency over FCN;
PSPNet \cite{zhao2016pyramid} and DeepLab \cite{chen2016deeplab} are introduced to
improve the prediction accuracy.

The proposed network model is also influenced by a few multi-task models, such as
MultiNet \cite{Teichmann2016}, StaffNet \cite{brahmbhatt2017stuffnet} and \cite{huval2015empirical}.
In our observations, the most closely related work to the proposed network model are MultiNet 
\cite{Teichmann2016} and StuffNet \cite{brahmbhatt2017stuffnet}, 
which predict the locations and semantic segmentation results with a shared convolutional architecture.
While MultiNet is the previous state-of-the-art multi-task model for 
real-time simultaneous object detection and segmentation, 
which is only enabled with single class detection and segmentation.
However, the proposed model can be seen as an improvement 
over the previous state-of-the-art multi-task convolutional network because it improves the Faster-RCNN-based
object detection framework with SSD-based network model and has the additional capability of predicting the distance 
for each detected object. It also improves the FCN-based segmentation framework with the 
pyramid pooling module introduced into PSPNet for the more precise prediction of semantic labels.
The following section describes the network design in detail.

\section{Details on Network Design}
\label{sec:network_design}

This section describes the proposed multi-task network architecture. The network contains
an encoder for feature extraction and two decoder branches for object detection, depth estimation and 
pixel-level semantic segmentation.
Both decoders use multi-level feature maps from the residual network based encoder.
They also use multiple convolutional layers for feature decoding.
As stated in Section \ref{sec:training}, the training strategy was optimized to
preserve both the instance-level features and pixel-level semantic features of the hierarchical network design.

\subsection{Shared Convolutional Architecture}

The design of the shared convolutional architecture enables efficient feature extractions for any  
street scene image that is provided as input.
The residual network \cite{He2015a} is used to perform feature extraction tasks.
More specifically, a $7\times{}7$ convolution along with first four residual blocks that were pretrained to perform an
image classification task upon ImageNet dataset \cite{ILSVRC15} are 
used as the shared convolutional architecture for image feature encoding.

In the case of resent-50, which was used in the proposed network model, 
there are 3/4/6/3 residual units in 1/2/3/4-th residual blocks, respectively.
Within each residual block, 
the resolution of the image features is reduced to half of the previous block size in terms of both height and width. 
The number of channels was extended to twice that of the previous block.
As the weights in network model were previously trained to predict image category, 
residual blocks that are close to the input layers tend to extract low-level features and preserve 
the location information of objects in the input image. 
The residual blocks that are close to the output layers tend to predict categorical features of the image
and contain less object location information.

At the training stage, gradient updates within the shared convolutional layers are influenced by both
detection and segmentation labels.
We proposed two training strategies to preserve both the instance-level features and 
the pixel-level semantic features within the shared convolutional architecture.
Details are described in Section \ref{sec:training}.

\subsection{Joint Object Detection and Depth Estimation}

We consider estimating the depth to be an additional component of the object localization task, along with
predicting the x- and y-axis location on the input image.

Following SSD \cite{Redmon2016}, we predict anchor boxes from the concatenated outputs of  
the multi-scale feature encoding layers (i.e., \texttt{res3} and \texttt{res4}) and 
convolutional architectures based on decoding layers, which are typically 4 stacked convolutional units.
Each unit contains a $1\times{}1$ convolution layer and a 3x3 convolution layer, 
with ReLU-based activation following each convolution layer.
For each anchor box, multiple aspect ratios are applied to each of the feature layers
to predict class labels, bounding boxes, and depth labels.
The design of the network enables the prediction of object detection with depth estimation that is based on hierarchical encoding layers.
The network benefits from details about the location of low-level encoding layers
and the instance-level object categorical information from higher level encoding layers.
Two branches are appended after each internal feature map, 
with one being the object categorical prediction branch that estimates the 
probabilities of each object category $c_1, c_2, ..., c_p$ per anchor box.
The other branch predicts the object center, size and depth information (a.k.a. $x,y,w,h,d$).
Detailed parameter settings are described in Figure \ref{fig:network_model}.

\textbf{Comparison with SSD framework} 
There are two major differences between the proposed detection branch and the original SSD \cite{Redmon2016}:
First, for each anchor box, along with the prediction of the object’s center and size, we add
an additional component that predicts depth from input feature maps.
Second, we replace the VGG-16-based feature encoding framework with ResNet-50, 
which achieved higher image classification accuracy with $\times{}5$ less parameters.


\subsection{Semantic Segmentation}

\textbf{Adaptively reduced number of outputs in the global prior and local feature maps.}
The design of the feature decoding layers is influenced by three existing frameworks: 
FCN \cite{Long2015}, SegNet \cite{badrinarayanan2015segnet} and PSPNet \cite{zhao2016pyramid}.
Specifically, to extract low-level feature maps, 
we follow the upsampling approach introduced in FCN \cite{Long2015} 
and use learnable upsampling filters by applying a deconvolution operator on the encoded low-level features. 
We also initialize the kernels with bilinear sampling weights.
According to SegNet \cite{badrinarayanan2015segnet}, 
neither bias is used in the convolution-based feature decoding layers. 
The ReLU non-linearity is followed by the convolution layers, 
which slightly increases the convergence rate at an early stage of training and
reduces the memory footprint at the inference step.
Additionally, inspired by the pyramid pooling module from PSPNet \cite{zhao2016pyramid}, 
a subsample from multiple scales of global features,
and upsamples followed by concatenating them into same scale, 
we use fewer outputs in the multi-scale pooling as a global prior.

More specifically, 
we reduced the number of outputs in both higher levels of the global prior and lower levels of local feature maps,
where the global prior is provided by higher level feature maps (i.e. $\mathtt{res5}$) via multi-scale average pooling. 
Typically, we use output sizes of 512/256/128 and stride sizes of 1/2/4, respectively, 
while local feature maps are upsampled from lower level shared convolution layers 
$\mathtt{res3}$ and $\mathtt{res4}$ with output size 128 and 256 respectively.
Before concatenating low-level feature maps into the adaptively reduced pyramid pooling module, 
it is preferred to process the low-level feature map with a $3\times{}3$ convolution layer.
The structure of the segmentation network is shown in Figure \ref{fig:network_model}.


\textbf{Reduce the output size of the segmentation task.}
Inspired by ENet \cite{Paszke2016} and SegNet \cite{badrinarayanan2015segnet} implementations, 
we reduced the output size of the segmentation task to be 4 times smaller in terms of both height and width.
Performing pixel-wise softmax at the input image scale would 
consume a large computational runtime and memory footprint,
which is unnecessary and less efficient.
As reported in the journal Cityscapes \cite{cordts2016cityscapes},
which subsampled the class-level semantic segmentation ground-truth by 2 / 4 / 8,
the IoU accuracy is decreased by 2.8 / 4.8 / 9.3 percent, respectively.
In our experiment, $1024\times{}512$ is the input size. 
By decreasing the segmentation output size by 4-fold in terms of both height and width,
the computation of softmax is reduced from 524,288 to 32,768 pixels, which is 16 times smaller.


\section{Supervised Training}
\label{sec:details}

In this section, we describe the implementation of details that have a great impact on the robustness and efficiency of the network model.

\subsection{Data Augmentation}

As stated in the SSD \cite{Redmon2016} article, data augmentation is critical at the training stage.
In contrast to the SSD design, which performs only the object detection task, the data augmentation used 
in the proposed network model derives depth ground-truth and segmentation results from a set of augmentation
parameters.

\textbf{Perform random flip, rotation, resizing for input augmentation.}
At the training stage, a random mirror and its random resizing between 0.5 and 2 is adopted.
The uniform random rotation is applied to each input image, with angle variations from -5  to 5 degrees, 
by simulating geo-vehicle bumping on the road. 
Since pure rotation does not result in scale changes of the augmented image, 
the provided depth value for each bounding box is untouched in case of rotation.
However, in the case of augmenting the input image with resizing, 
e.g., by a factor of $s_x$ and $s_y$, 
the ground-truth distance is scaled by $\sqrt{s_x s_y}$.

To provide sufficient detail for detecting small objects in a scene, 
we use the raw image as the input and resize it via bilinear interpolation.

\textbf{Eliminate overwhelming small object speculation.}
In our preliminary experiment, without limiting the distance range of objects for detection at training stage, 
the network falls to detecting many imaginary traffic signs. 
We hypothesize the reason to be that there are many small ground-truth subjects of the ground-truth.
Also, as stated in SSD \cite{Redmon2016}, small objects are rarely correctly detected, which may lead to
0 mAP for very small objects at inference stage.
Therefore, ground-truth bounding-boxes with area sizes below 100 are ignored
when generating augmented training dataset.


\begin{figure*} 
   \centering
   \includegraphics[width=6.8in]{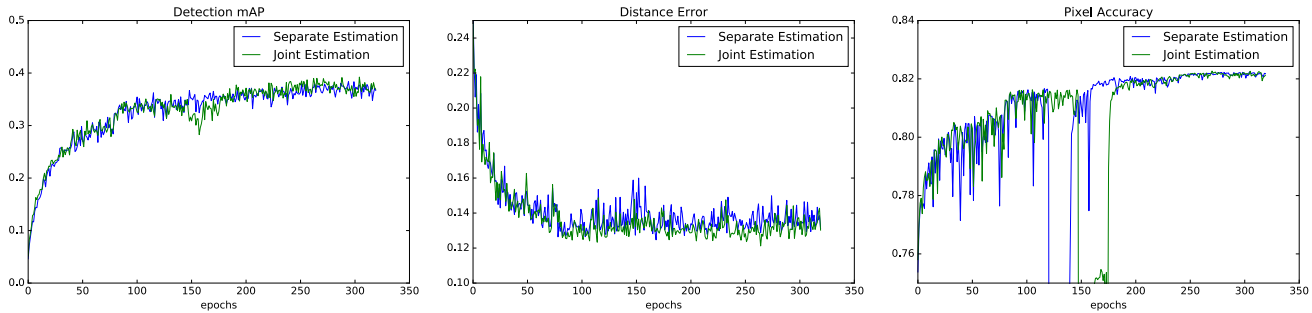} 
   \caption{Comparison of separate and joint task training of DSPNet on the Cityscapes dataset. 
   All figures shown above are evaluated using the Cityscapes validation set for each epoch during training, 
   where the detection mAP indicates the detection accuracy of the resulting bounding boxes; 
   the distance error indicates the accuracy of distance prediction across all object classes; 
   and the pixel accuracy curve indicates changes in the pixel-level prediction accuracy in semantic segmentation tasks.}
   \label{fig:curves}
\end{figure*}

\subsection{Training}
\label{sec:training}

Formally, we define a multi-task loss for each input image as $L=L_{cls}+L_{reg}+L_{seg}$.
The classification loss $L_{cls}$ is identical to the SSD\cite{Redmon2016}.
We extend the localization loss in SSD $L_{loc}$  with additional depth regression loss into
$L_{reg}$.

\textbf{Scale gradient for balancing sub-task losses.}
Since both detections and segmentations have impact on the gradient evolution of 
the shared convolutional architecture, to balance the overall accuracy of both tasks, we 
add an additional weight $w_{seg}$ to the segmentation loss $L_{seg}$ to adjust for the impact between 
the segmentation branch and detection branch. 
In our experiments with the Cityscapes dataset, where we only use $1024\times{}512$ as the input image size,  
the output layer for detection generated 12,264 prior boxes.
The output segmentation result size is set to $256\times{}128$ pixels.
The number of object classes for detection is 10, 
and the number of classes for semantic segmentation is 19.
The resulting gradient scale weight $w_{seg}$ is set to 4 for DSPNet.

\textbf{Block propagation of the gradient from the segmentation task to low-level residual units.}
To improve the capability of performing detection and segmentation simultaneously
while preventing the prediction of multiple instances with a single bounding box, 
the propagation of gradient from softmax loss in the segmentation task 
towards $\mathtt{res3}$ and $\mathtt{res4}$ residual units has been blocked in the the training stage. 
This strategy causes the proposed network to preserve instance-level details in 
lower level residual units (i.e., $\mathtt{res3}$ and $\mathtt{res4}$) and further processes the feature maps into 
higher level residual units (i.e., $\mathtt{res5}$).


\textbf{Optimization}
We use initial learning rate of 0.0005 with a multi-factor learning rate scheduler for 
gradient update.
The learning rate decreases at the rate of 0.5 at epoch 80/160/240, respectively.
The training is terminated after 320 epochs, and the momentum for each gradient update is set to 0.9.
Additionally, due to limited size of GPU memory, 
we use the batch size of 2 for training the network with $1024\times{}512$ as the input image size.
A comparison of the validation accuracy cruves between separate and joint training the proposed network model is
shown in Figure \ref{fig:curves}. 


%


\section{Experimental Results}
\label{sec:results}

The section describes our experimental results based on the Cityscapes dataset \cite{cordts2016cityscapes}.
All experiments has been performed on NVIDIA GeForce GTX-1080, and all timing results are median filtered over 500 trails.

\begin{table*}
\begin{center}
\scalebox{1}{
\begin{tabular}{c|c c c c c c c c c c c c c c c c c c c c}
\Xhline{2\arrayrulewidth}
Method & person & rider & car & truck & bus & train & mbike & bike & mAP \\ \hline \hline
SSD /w depth (baseline) & 36.3 & 37.2 & 60.2 & 26.6 & 47.5 & 28.8 & 26.9 & 30.1 & 36.7 \\ \hline \hline
StuffNet \cite{brahmbhatt2017stuffnet} & 17.1 & 26.7 & 39.3 & 10.6 & 28.8 & 12.7 & 0.0 & 19.1 & 19.3 \\ \hline
DSPNet & 34.9 & 37.7 & 59.1 & 29.4 & 49.3 & 30.4 & 24.6 & 30.0 & 36.9 \\
                                                          
\Xhline{2\arrayrulewidth}
\end{tabular}
}
\end{center}
\caption{Detection results for each class on the Cityscapes dataset.}
\label{tab:det_class}
\end{table*}%

\subsection{Dataset}

The Cityscapes dataset provides both fine and coarse level annotations with up to 34 object categories 
at the pixel level. To prevent ambiguity in predicting scene segmentation task, only 19
of the 34 object categories are assigned to be predicted in a scene segmentation task, 
and 8 them are assigned as instance objects to be detected in an object detection 
or instance-level segmentation task.
For the fine-level annotated set of the complete dataset, 
it provides 2975/500/1525 images for training/validation/testing, respectively.
Additionally, the dataset provides disparity maps that are precomputed from calibrated stereo camera systems.

\subsection{Distance Ground-truth Generation}

Because disparity and distance from the cameras are inversely related, 
the distance ground-truth is generated from the disparity map by computing

\begin{equation}
\label{eq:dist_gt}
D_{gt} = b*f/d,
\end{equation}

\noindent where $D$ is the distance between the camera and an object in real world, $b$ is the base offset
(i.e., the distance between cameras), and $f$ is the focal length of the camera with $d$ being disparity.

Here, we assume the depth of each bounding box is similar. As we illustrated below, 
the estimated distance should close to average of all the pixels of the target.
The ground-truth distance label for each bounding box is computed from disparity map provided in the Cityscapes dataset.
To improve the robustness of the process, we perform median filtering of valid disparity values within the bounding box.

\subsection{Prediction Accuracy Analysis}

\begin{figure*} 
\begin{center}
   \includegraphics[width=6.8in]{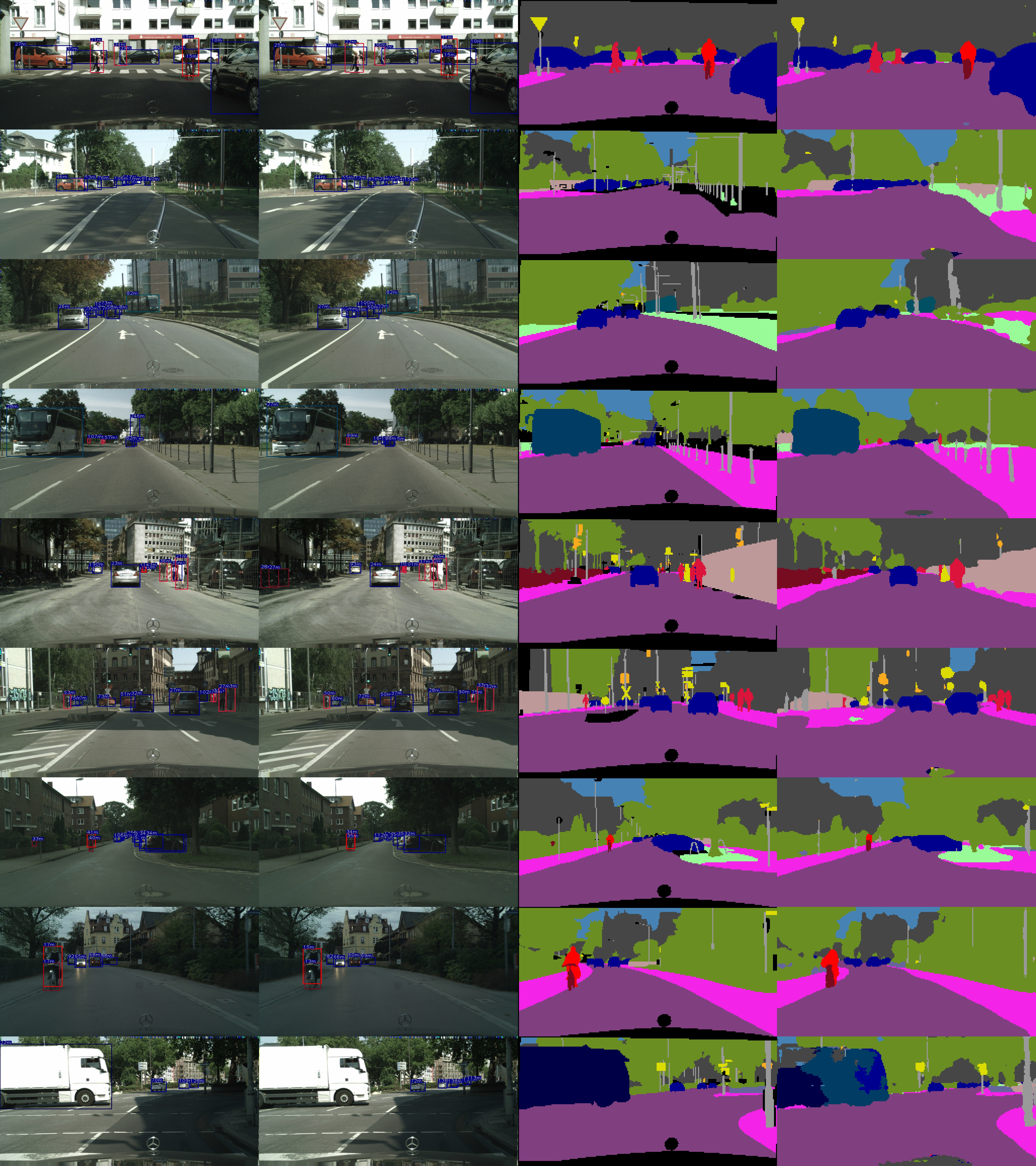} 
\end{center}
    \begin{subfigure}[b]{0.24\textwidth}
        \caption{Detection GT}\label{fig:results_a}
    \end{subfigure}   
    \begin{subfigure}[b]{0.24\textwidth}\label{fig:results_b}
        \caption{Detection Result}
    \end{subfigure}   
    \begin{subfigure}[b]{0.24\textwidth}\label{fig:results_c}
        \caption{Segmentation GT}
    \end{subfigure}   
    \begin{subfigure}[b]{0.24\textwidth}\label{fig:results_d}
        \caption{Segmentation Result}
    \end{subfigure}   
\caption{Experimental results of the Cityscapes validation set. Specifically, 
ground-truth detection of the bounding boxes are shown in column (a); 
our detection results are shown in column (b);
segmentation ground-truth images are shown in column (c);
and our segmentation results are shown in column (d). 
The results in the last row failed to predict the trunk as expected.}
\label{fig:results}
\end{figure*}

In this section, we analyze the prediction accuracy of the network among 
accuracy of object detection, depth prediction and segmentation tasks.
Several experimental results based on the validation set are shown in Figure \ref{fig:results}.

\subsubsection{Detection}

Since we found no publication reporting
bounding box level prediction results on the Cityscapes dataset,
the evaluations of the robustness and effectiveness of the proposed model are based on 
our baseline model without a joint estimation of semantic segmentation, 
and an existing implementation of StuffNet.
Comparisons with the existing approaches are based on the bounding box level detection results for each class as shown in Table \ref{tab:det_class}.
Note that ground-truth bounding boxes with area sizes smaller than 100 pixels are ignored
both at training stage and evalutation stage for all experiments.
As the Cityscapes dataset is annotated with sufficient details,
the proposed network model significantly out-perform StuffNet\cite{brahmbhatt2017stuffnet}
on both of the detection and segmentation tasks.
And the proposed multi-task network also slightly out-perform both of the single-task network models,
which have been denoted as SSD /w depth and Reduced PSPNet.


\subsubsection{Depth Estimation}

\begin{figure}[H]
\begin{center}
\includegraphics[width=3.2in]{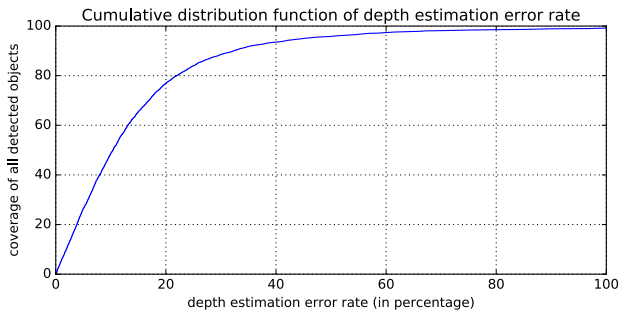}
\caption{Cumulative distribution function (CDF) of distance error rates estimated with the proposed network
on the Cityscapes validation set.}
\label{fig:distance_cdf}
\end{center}
\end{figure}

To further analyze the distribution of depth estimation error, we calculated the cumulative distribution function
of all the detected objects, and the resulting curvature is shown in Figure \ref{fig:distance_cdf}.
It shows that the depth for 78\% of the detected objects are
estimated within 20\% relative estimation error.

\vspace*{-.2in} 

\begin{figure}[H]
\begin{center}
\includegraphics[width=3.2in]{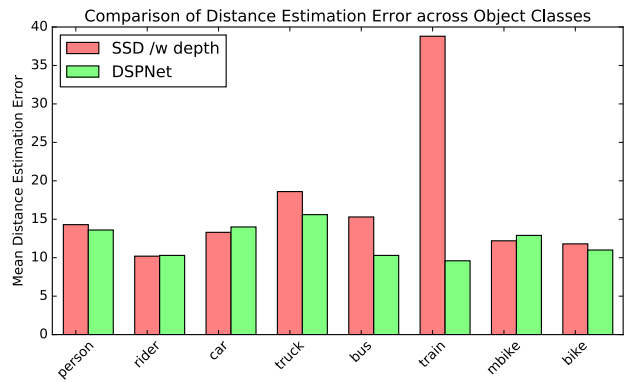}
\caption{Comparison of depth estimation error between SSD /w depth and DSPNet across object classes.
As buses and trains are similar in appearance,
DSPNet out-performs the single-task model in estimating distance to a train.}
\label{fig:dist_error_barchart}
\end{center}
\end{figure}

For each detected object, we evaluate the error between estimated distance and ground-truth 
distance from the disparity map. At this stage, the correctness of the bounding box regression and 
the object classification is ignored.
The distance error for each bounding box is measured as follows:

\begin{equation}
\label{eq:dist_error}
error = \frac{|D_{est}-D_{gt}|}{D_{gt}},
\end{equation}

where $D_{est}$ is the estimated distance of a single bounding box
and $D_{gt}$ is defined in Eq. (\ref{eq:dist_gt}) as being the ground-truth distance calculated from a disparity map.
Additionally, the depth estimation error for each object class has been evaluated
as shown in Figure \ref{fig:dist_error_barchart}.


\begin{table*}
\begin{center}
\scalebox{0.68}{
\begin{tabular}{c|c c c c c c c c c c c c c c c c c c c c}
\Xhline{2\arrayrulewidth}
Method & road & swalk & build. & wall & fence & pole & tlight & sign & veg. & terrain & sky & person & rider & car & truck & bus & train & mbike & bike & mIoU \\ \hline 
\hline
PSPNet \cite{zhao2016pyramid} & 98.6 & 86.2 & 92.9 & 50.8 & 58.8 & 64.0 & 75.6 & 79.0 & 93.4 & 72.3 & 95.4 & 86.5 & 71.3 & 95.9 & 68.2 & 79.5 & 73.8 & 69.5 & 77.2 & 78.4 \\ \hline
DeepLab \cite{chen2016deeplab} & 97.9 & 81.3 & 90.3 & 48.8 & 47.4 & 49.6 & 57.9 & 67.3 & 91.9 & 69.4 & 94.2 & 79.8 & 59.8 & 93.7 & 56.5 & 67.5 & 57.5 & 57.7 & 68.8 & 70.4 \\ \hline
FCN \cite{Long2015} & 97.4 & 78.4 & 89.2 & 34.9 & 44.2 & 47.4 & 60.1 & 65.0 & 91.4 & 69.3 & 93.9 & 77.1 & 51.4 & 92.6 & 35.3 & 48.6 & 46.5 & 51.6 & 66.8 & 65.3 \\ \hline
Reduced PSPNet \dag & 96.8 & 75.7 & 87.4 & 40.7 & 40.9 & 40.1 & 45.3 & 56.5 & 88.3 & 56.4 & 90.9 & 65.0 & 41.7 & 89.3 & 65.1 & 73.9 & 65.7 & 40.6 & 63.5 & 64.4 \\ \hline
\hline
StuffNet \dag & 96.6 & 74.2 & 87.4 & 36.0 & 43.9 & 33.8 & 40.0 & 54.1 & 87.9 & 55.7 & 90.4 & 66.4 & 40.6 & 89.7 & 41.1 & 64.9 & 53.0 & 29.4 & 63.6 & 60.5 \\ \hline
DSPNet \dag & 96.5 & 75.2 & 87.7 & 43.7 & 41.5 & 37.6 & 43.6 & 56.6 & 87.7 & 56.2 & 90.4 & 67.3 & 45.5 & 89.2 & 60.4 & 79.2 & 59.9 & 50.7 & 63.6 & 64.9 \\
\Xhline{2\arrayrulewidth}
\end{tabular}
}
\end{center}
\caption{Segmentation results for each object class on Cityscapes dataset.
The first four rows are results from single task segmentation network models,
while the last two rows are results from multi-task models.
Since the testing set ground-truth in the Cityscapes dataset is only available on the evaluation server,
the rows labeled with $\dag$ are results based on the validation set,
while others are based on the testing set.}
\label{tab:seg_classes}
\end{table*}%

\begin{table*}
\begin{center}
\scalebox{.95}{
\begin{tabular}{c|c c c c c}
\Xhline{2\arrayrulewidth}
Method & Memory Usage & Param. Size (FP32) & Runtime \\ \hline \hline
SSD w/ depth & 646 MiB &  130.5 MiB &  61.5 ms \\ 
Reduced PSPNet & 722 MiB &  122.8 MiB &  55.0 ms \\ \hline
SSD w/ depth + Reduced PSPNet & 1368 MiB &  253.2 MiB &  116.5 ms \\ \hline \hline 
StuffNet \cite{brahmbhatt2017stuffnet} &  2682 MiB &  598.8 MiB &  144.2 ms \\ \hline
DSPNet & 836 MiB & 144.4 MiB & 71.3 ms \\
\Xhline{2\arrayrulewidth}
\end{tabular}
}
\end{center}
\caption{Computational efficiency and memory usage comparison among the combination of single tasks, StuffNet and the proposed network model.}
\label{tab:efficiency}
\end{table*}%

\vspace*{-.2in}

\subsubsection{Segmentation}

For the segmentation task, we followed PSPNet \cite{zhao2016pyramid} to use $1024\times{}512$
as the input image size
and performed upsampling of the probability map of segmentation results via bilinear sampling
and generated standard segmentation result for direct comparison with state-of-the-art results.
The overall segmentation performance is described in Table \ref{tab:seg_overall}, 
and a comparison with the segmentation results for each object class
is shown in Table \ref{tab:seg_classes}.
As we can see from Table \ref{tab:seg_overall},
the proposed multi-task model achieves higher iIoU comparing to the single-task segmentation network.
We hypothesize that it is the contribution of the additional detection network
that makes the multi-task model more robust.


To directly evaluate the down-sampled segmentation results, 
we also computed class-level IoU values between predicted segmentation results with 
subsampled ground-truth segmentation annotation, 
where we use nearest neighborhood for sub-sampling segmentation label.
We achieved 64.9\% class-level IoU accuracy
upon validation set of the Cityscapes dataset after training the network with 320 epochs.


\begin{table}[H]
\begin{center}
\scalebox{0.85}{
\begin{tabular}{c|c c c c}
\Xhline{2\arrayrulewidth}
Method & IoU cls. & iIoU cls. & IoU cat. & iIoU cat. \\ \hline \hline
FCN \cite{Long2015} & 65.3 & 41.7 & 85.7 & 70.1 \\ \hline
DeepLab \cite{chen2016deeplab} & 70.4 & 42.6 & 86.4 & 67.7 \\ \hline
SegNet \cite{badrinarayanan2015segnet} \dag &  56.1 & 34.2 & 79.8 & 66.4 \\ \hline
ENet \cite{Paszke2016} \dag &  58.3 & 34.4 & 80.4 & 64.0 \\ \hline 
Reduced PSPNet \dag & 64.4 & 36.6 & 80.8 & 57.1 \\ \hline \hline
StuffNet \cite{brahmbhatt2017stuffnet} \dag & 60.5 & 37.8 & 80.3 & 63.0 \\ \hline
DSPNet \dag & 64.9 & 40.9 & 80.8 & 60.3 \\
\Xhline{2\arrayrulewidth}
\end{tabular}
}
\end{center}
\caption{Segmentation performance comparison based on the Cityscapes dataset. 
Items labeled with \dag{} are results based on validation set, others are based on testing set.}
\label{tab:seg_overall}
\end{table}%


\subsection{Runtime Requirement Analysis}


As stated above, the computational efficiency at the inference stage is 
one of the most important considerations of our overall network design. 
Therefore, we further analyze how many resources are allocated for 
each task and try to eliminate timing-consuming and less effective layers in the network.
A comparison of runtime requirements with state-of-the-art models is shown in Table \ref{tab:efficiency}.
It shows that the joint trained DSPNet is more efficient
than a simple combination of the single-task models.

\begin{figure}[H]
\begin{center}
        \includegraphics[width=3.2in]{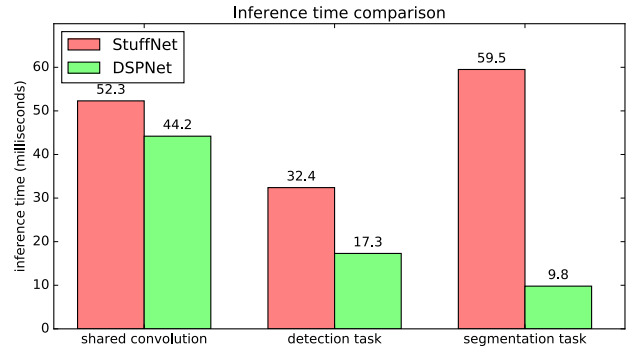}
\caption{Inference time comparison of individual tasks between StuffNet and DSPNet.
}
\label{fig:inference_time_pie_chart}
\end{center}
\end{figure}

Here, we partition the network into three sections: 
feature encoding layers,  detection task branch and segmentation task branch.
As shown in Figure \ref{fig:inference_time_pie_chart}, the 
most time-consuming section of the proposed network is the detection network, 
which predicts bounding boxes and distance from multi-scale feature maps.

\vspace*{-.2in}


\section{Conclusion and Future Works}
\label{sec:conclusion}

As shown in the previous section, the proposed network model is capable of 
simultaneous object detection, depth estimation and semantic segmentation 
with a shared convolutional architecture and is proven to be efficient at inference time. 
However, many other efforts have been made to improve efficiency of the prediction, 
such as increasing the sparsity of the receptive fields and quantizing the inference computation.
These techniques would help the proposed network to be more efficient.
In summary, the proposed multi-task network model is suitable to be ported for embedded systems, 
thus helps drivers avoid frontal collision cases; alternatively, it can be used as a visual perception module
in an automated driving system.



\clearpage

\clearpage

{\small
\bibliographystyle{ieee}
\bibliography{cvpapers}
}

\clearpage


\end{document}